# Identifying Pairs in Simulated Bio-Medical Time-Series

Uri Kartoun, *Member, IEEE*

*Abstract*—The paper presents a time-series-based classification approach to identify similarities in pairs of simulated human-generated patterns. An example for a pattern is a time-series representing a heart rate during a specific time-range, wherein the time-series is a sequence of data points that represent the changes in the heart rate values. A bio-medical simulator system was developed to acquire a collection of 7,871 price patterns of financial instruments. The financial instruments traded in real-time on three American stock exchanges[1], *NASDAQ*, *NYSE*, and *AMEX*, simulate bio-medical measurements. The system simulates a human in which each price pattern represents one bio-medical sensor. Data provided during trading hours from the stock exchanges allowed real-time classification. Classification is based on new machine learning techniques: *self-labeling*, which allows the application of supervised learning methods on unlabeled time-series and *similarity ranking*, which applied on a decision tree learning algorithm to classify time-series regardless of type and quantity.

## I. INTRODUCTION

Real-time sensing using wearable technologies includes variety of measurements such as heart rate, brain activity, and hydration levels to name a few. Examples for human-generated sensing and monitoring systems provided by [1] include: [2]-[6]. [7] provide a comprehensive review of variety of systems and sensing devices, including bio-chemical sensors for collecting fluid [8][9], accelerometers to identify epilepsy seizures [10], accelerometers to monitor *COPD* [11], photoplethysmographic bio-sensors to monitor cardiovascular activity [12]-[14], and pulse oximeters to monitor oxygen saturation [15]. Additional examples include using *Complex Event Processing* with bio-sensors, *RFID*, and accelerometers to evaluate health-related events [16], *Zigbee* modules to monitor temperature and heart rate [17], *Electromyography (EMG)*, *Electrodermal activity (EDA)*, and pressure respiration sensors to monitor mental stress [18], Electrocardiography (*ECG*) recording to measure respiration rate [19]. An example for a continuous assessment of *EDA* outside of a laboratory setting is provided in [20].

*NeuroGlasses* is a wearable physiological signal monitoring system [21]—the system acquires health-related signals by using *OCZ NIA*, a head-band with one neurosensory; a location-aware signal correction and a feature-based re-construction approach were proposed to compensate signal distortion due to the variation of sensor location and other noises. [22] describe an implementation of decision tree and artificial neural network algorithms to classify activities such as walking, running, and cycling; a data collection system was developed and provided measurements to 18 different signals, such as, altitude, audio, chest and wrist acceleration, *EKG*, humidity, light intensity, heart rate, respiratory effort, and skin temperature. A related study focused on human movement monitoring introduces a method for temporal parameter extraction. The method, *Hidden Markov Event Model (HMEM)*, can be adapted to new movements and new sensors and was validated on a walking data-set by using sensors such as triaxial accelerometer and biaxial gyroscope [23].

One limitation of current wearable monitoring and prediction systems is that they are designed to acquire information using specific one or several sensors to address a certain medical condition. The ability to add other sensors to an existing system of the same type or other types is limited and usually requires re-designing the systems. Another limitation is the complexity of fusioning the measurements acquired by multiple sensors. Fusioning is even more challenging as technology is advancing rapidly and new types of sensors are becoming available, e.g., *Q-Sensor* [24]. Systems that were already developed to fusion several existing sensors are limited in scaling up with new types of sensors. Further, existing systems are typically capable of analyzing sensing information acquired by only a few sensors. In the era of social networks and big data it is expected that the ability to measure human-generated signals both quantitatively and qualitatively will significantly improve over the next few years. The improved ability will allow the systems to provide higher quality of monitoring and better prediction accuracy of undesired medical situations (e.g., an epilepsy seizure, a heart attack, or an occurrence of a fall).

Another limitation of existing systems is their relatively simple computational mechanisms which are usually based on observing events, i.e., a certain sensor measurement value is

---

above or below a pre-defined threshold, or they are based on analyzing short patterns of measurements, usually for one sensor at a time.

Another limitation relates to discovering associations between the measurements, i.e. to classify measurements from different sensors, for example, to identify a pair comprised of similarly behaving time-series representing skin conductivity and heart rate. Such ability has not yet been sufficiently addressed.

This paper describes a time-series based classification approach to identify similarities in patterns. An example of a pattern would be a time-series of values measured by a certain sensor during a specific time period, wherein the time-series is a sequence of data points that represent the change in the sensor value (e.g., measured every second). The approach is based on applying new machine learning methods: *self-labeling* and *similarity ranking* that allow classifying of time-series regardless of type and quantity. The paper describes further the implementation of a simulation system which is capable of acquiring a large collection of time-series and to identify similarities among the time-series. The system is based on three stages of operation: *sampling*, *classification*, and *tracking*. To evaluate the system's capabilities, a collection of 7,871 price patterns of financial instruments traded in real-time on three American stock exchanges, *NASDAQ*, *NYSE*, and *AMEX*, was used. No existing bio-medical-based system is capable of sampling and analyzing such a large amount of time-series representing sensors in real-time, as it is not yet possible to connect more than a few sensors to a one person. The simulation system could then be observed as a human in which the financial instruments are considered as simulated bio-medical measurements.

The structure of the paper is as follows: after introducing the bio-medical simulator in Section II, the time-series classification methods are detailed in Section III. Section IV presents an experiment to validate the classification accuracy. Discussion and conclusions are provided in Sections V and VI, respectively.

## II. A Bio-Medical Simulator for Identifying Pairs

### A. Overview

To simulate the measurements of a large collection of bio-medical sensors, a classification system and several computational methods were developed. The system comprised of three operational stages—1) *Sampling* - the procedure of acquiring time-series over a pre-defined time range, 2) *Classification* - the procedure of identifying similarities among the time-series, and 3) *Tracking* - the procedure of having a user or an automatic monitoring process the ability to observe in real-time pairs of time-series that were classified as similarly behaving during the sampling and classification stages and act on them.

### B. Sampling

The sampling stage comprised of sampling $m$ time-series sampled several times from time $T_1$ to time $T_2$ (Fig. 1). Sampling of time-series includes a first sampling of the values of $m$ time-series measured at time $T_1$, several more samples of the values of the $m$ time-series (e.g., every ten seconds), and a last sampling of the values of the $m$ time-series measured at time $T_2$. The number of samples, $n$, is determined in advance as well as the interval between samples. There is no limitation, however, to have an identical interval value between the samples. When sampling ends, the vectors of values (e.g., prices representing financial instruments) are as represented in (1).

### C. Classification

The classification stage identifies pairs of similarly behaving financial instruments. The steps are based on applying several methods, including self-labeling, decision tree learning, and similarity ranking. When the classification stage ends, a sub-set of the original time-series is stored in a database. The sub-set contains pairs of time-series that were found as similarly behaving. The methods are described in greater detail in Section III.

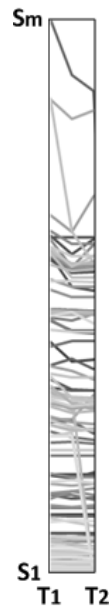

Fig. 1. An example for sampling time-series.

### D. Tracking

Once the *Sampling* and *Classification* procedures end, the identified pairs of financial instruments are presented on the user interface. Next to each financial instrument presented in real-time several parameters including its current price (given in U.S. dollars), the change in price since tracking started (given in percent) and its sector (Fig. 2). A sector could have one of the following values: 1) Services, 2) Healthcare, 3) Utilities, 4) Financial, 5) Consumer Goods, 6) Basic Materials, 7) Conglomerates, 8), Industrial Goods, and 9) Technology. The user has the option to present only pairs in which in each pair the financial instruments belong to the same sector. The financial instruments are visualized in real-time to let the user observe and monitor the acquired data.

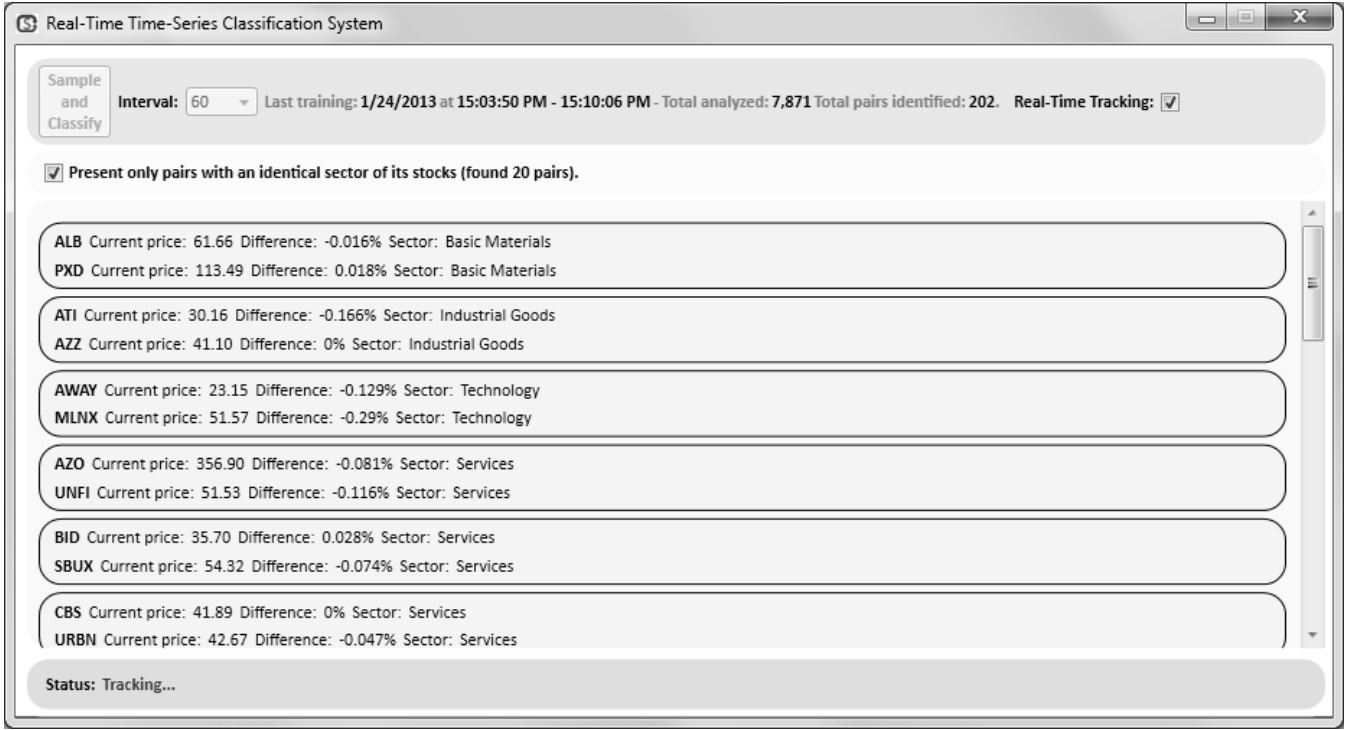

Fig. 2. User interface - observing 202 pairs (January 24, 2013).

## III. TIME-SERIES CLASSIFICATION

### A. Self-Labeling

Assume $S_1, S_2, S_i, ... S_m$ are $m$ financial instruments considered for classification during a trading time range that includes $n$ time-steps (e.g., one time-step equals 10 seconds). Each financial instrument $S_i$ is associated with a vector of prices in which vector of prices is denoted as $S(P)$. For a financial instrument $S_i$ the vector of prices, is presented as follows:

$$S_1(P_{t_1}, P_{t_2}, P_{t_3}, P_{t_j}, ... P_{t_n})$$
$$...$$
$$S_m(P_{t_1}, P_{t_2}, P_{t_3}, P_{t_j}, ... P_{t_n}) \quad (1)$$

For all financial instruments, generate vectors representing the change in price for every two subsequent time-steps:

$$S_1(\frac{P_{t_2}}{P_{t_1}}-1, \frac{P_{t_3}}{P_{t_2}}-1, \frac{P_{t_4}}{P_{t_3}}-1, ... \frac{P_{t_n}}{P_{t_{n-1}}}-1)$$
$$...$$
$$S_m(\frac{P_{t_2}}{P_{t_1}}-1, \frac{P_{t_3}}{P_{t_2}}-1, \frac{P_{t_4}}{P_{t_3}}-1, ... \frac{P_{t_n}}{P_{t_{n-1}}}-1) \quad (2)$$

To simplify the representation of (2) it is presented as:

$$S_1[C_1, C_2, C_3, ... C_n]$$
$$...$$
$$S_m[C_1, C_2, C_3, ... C_n] \quad (3)$$

In the classification problem considered here no labels are available for the time-series and there is no information on how to refer to a set of values associated with a certain time-series. As such, a numerical value representing each time-series is generated and assigned as the label of the time-series. The numerical value label denoted as $LS_i$ is calculated for each time-series:

$$LS_1 = \sum_{l=1}^{n} S_1(C_l)$$
$$... \quad (4)$$
$$LS_m = \sum_{l=1}^{n} S_m(C_l)$$

The representation of self-labeling as shown in (4) facilitates the application of supervised learning methods on unlabeled data sets. This is achieved by providing a supervised learning classification algorithm with pairs of adjusted representations of original signals (3) and the adjusted representations' corresponding self-generated label (4).

### B. Decision Tree Learning

The patterns are stored and modified using a data preparation procedure as described through (1) - (4). The procedure described through (1) - (4) is applied by acting several tables stored in a database. Values according to (3) and their corresponding labels (4) are served as an input for a standard supervised learning algorithm. The supervised learning algorithm used is a decision tree learning algorithm. The algorithm is applied on the entire price patterns of all financial instruments considered once the sampling stage ends. 7,871 financial instruments are considered for classification including *AMEX*, *NYSE*, and *NASDAQ*. The total time range for classification depends on a time interval value set by the user through the user interface. A time interval, for example, equals to 10 seconds represents a sampling period of one minute considering that the number of samples is a pre-defined value equals to six ($n = 6$ in (3) and (4)) [2].

An example for a partial representation of a decision tree is shown in Fig. 3[3]. A decision tree is a data structure that consists of branches and leaves. Leaves (also denoted as "nodes") represent classifications, and branches represent conjunctions of features that lead to those classifications. Each node has a unique title to distinguish the node from other nodes that the tree is composed of. A node contains two or more records. Each record represents a financial instrument, its feature values (2) and its predictor value (4). The fewer records in a node (the minimum is two), the less this node varies, i.e., a node with fewer records is more likely to represent a better classification between the financial instruments that the node contains.

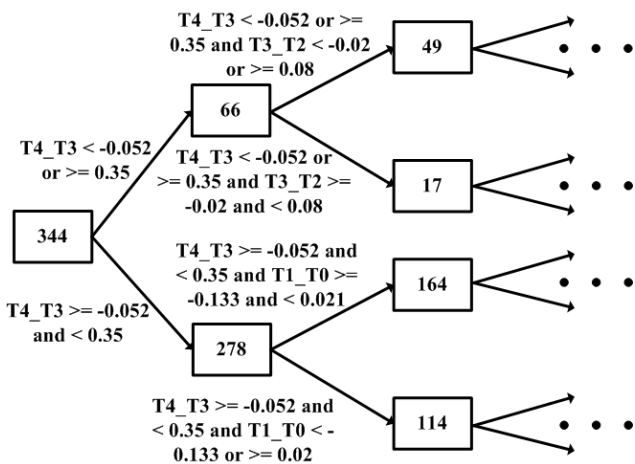

Fig. 3. A partial representation of a decision tree (total of 263 nodes, January 24 2013, 15:03:50 PM - 15:10:06 PM).

The number of nodes in a generated tree depends on the number of time-steps and the number of financial instruments considered. The classification accuracy of the algorithm depends on its input parameters. Parameters for a decision tree algorithm include complexity penalty, to control the growth of the decision tree, and minimum support, to determine the minimal number of leaf cases required to generate a split.

Setting the desired values for the decision tree algorithm parameters depends on the tradeoff between classification accuracy and computational speed. Classifying with perfect or close to perfect accuracy a large number of time-series may require an unfeasible processing time. To reduce the calculation time, the growth of the decision tree is controlled by increasing the complexity penalty level (this decreases the number of splits) and by increasing the level of minimum support. On one hand, controlling the growth of the tree improves computation performance. On the other hand, controlling the growth of the tree may affect classification accuracy.

When the sampling stage ends, a decision tree based classification is performed. For the amount of data considered here, a typical size for one decision tree is in the range of several-hundred nodes. The decision tree includes a main node that contains all financial instruments. The decision tree algorithm generates rules. The rules are based on values for the financial instruments (price change) for every two subsequent samples; as described through (1) - (4). Some nodes in the tree split to two sub-nodes, i.e., children, and other nodes do not. A split, if occurs, is based on the generated rules and separates a group of financial instruments to two smaller groups. For example, for the main node that consists of 344 financial instruments, two rules were generated[4]:

- *T4_T3 < -0.052 or >= 0.35*, and,
- *T4_T3 >= -0.052 and < 0.35*.

Similarly, other generated rules split nodes across the tree. The decision tree classification results for the sampling stage considered are stored in a database.

*C. Similarity Ranking*

To receive a similarity for a financial instrument comprising together a pair, a *similarity ranking* method is applied (Fig. 4). Consider a financial instrument—the financial instrument denoted as $S$ queries the decision tree data structure. Similarly behaving financial instruments are identified and sorted according to the number of occurrences at the nodes that also contain $S$. The financial instrument with the highest counter value is picked to be part of the pair. In case two or more financial instruments with an identical counter value were found then one of them is picked arbitrarily.

---

[2] Sampling period could be longer, depending on Internet communication latency and on potential data access delays.

[3] The figure presents a decision tree that includes 344 of the entire set of 7,871 time-series considered. The reason for that—time-series that do not contain all prices for the period considered are omitted from the analysis for that time range (typically six values). Additionally, the algorithm omits financial instruments with insignificant trading activity during the sampling.

[4] $T_n\_T_{n-1}$ for a financial instrument represents the difference given in percent between the price sampled at time $T_n$ to the price sampled at time $T_{n-1}$.

> Given a decision tree
>    Find all nodes that contain $S$
>       Find financial instruments in a node and increase by 1 a counter value associated with each financial instrument.
> Sort the financial instruments in a descending order according to the total counter value of a financial instrument.

Fig. 4. Similarity ranking method.

## IV. EXPERIMENTS

To evaluate the classification accuracy of the simulator, the sampling period followed by classification (January 24 2013, 15:03:50 PM - 15:10:06 PM) was observed. 202 pairs were identified as pairs in while in each pair the two financial instruments were found as similarly behaving (see examples in Fig. 5). The 20 similarly behaving pairs with the same sector for the two financial instruments in a pair along with the *Pearson Product-Moment Correlation Coefficient* values ($r$) for each pair are presented in Table A. The correlation of 90% of the pairs (18 out of 20) was positive indicating high classification accuracy. The average of $r$ for the 20 pairs was 0.63 (SD: 0.39).

TABLE A
SIMILARLY BEHAVING PAIRS AS OBSERVED DURING JANUARY 24 2013, 15:03:50 PM - 15:10:06 PM

| Pair | | Sector | $r$ |
|---|---|---|---|
| ALB | PXD | Basic Materials | 0.82 |
| ATI | AZZ | Industrial Goods | 0.85 |
| AWAY | MLNX | Technology | 0.86 |
| AZO | UNFI | Services | 0.96 |
| BID | SBUX | Services | 0.87 |
| CBS | URBN | Services | -0.26 |
| CDE | SEMG | Basic Materials | 0.97 |
| CSTR | DV | Services | 0.18 |
| DECK | SCSS | Consumer Goods | 0.95 |
| DRC | ROP | Industrial Goods | 0.63 |
| EPD | INT | Basic Materials | -0.40 |
| GCO | NFLX | Services | 0.76 |
| GEOY | NTES | Technology | 0.73 |
| HZO | LAD | Services | 0.87 |
| JBHT | FNGN | Services | 0.37 |
| KMP | CLB | Basic Materials | 0.46 |
| KSS | WYNN | Services | 0.82 |
| UA | DECK | Consumer Goods | 0.69 |
| URI | VSI | Services | 0.62 |
| WAG | ATHN | Services | 0.74 |

## V. DISCUSSION

Providing classifications for signals or time-series is well discussed in the literature; however, what is significant here as described through (1) - (4) is a *self-labeling* enhancement that facilitates the application of supervised learning methods on unlabeled data sets. Another significance facilitates classifying time-series—the *similarity ranking*, which applied on a decision tree learning algorithm to classify time-series regardless of type and quantity.

The methods described in the paper as applied on financial time-series could be implemented on a large collection of bio-medical sensors—the simulator presented in the paper allows

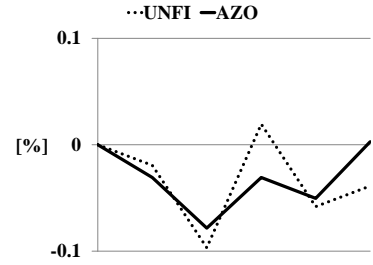

(a) UNFI and AZO ($r$=0.96).

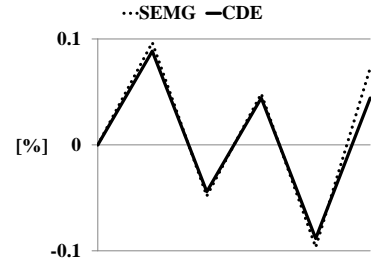

(b) SEMG and CDE ($r$=0.97).

Fig. 5. Examples for pairs with a high correlation (January 24 2013, 15:03:50 PM - 15:10:06 PM).

sampling and classifying unspecified number of time-series measurements. Additional financial readings, i.e., sensor readings, could be acquired without modifying the design and without having to develop additional software components. Another advantage of the methods is that they have the potential to reduce the complexity that involves in fusioning the measurements acquired by multiple sensors. The ability to fusion readings from multiple and unspecified number of sensors, including new types of sensors, has the potential to include additional sensors as part of an existing monitoring bio-medical system (e.g., *Q-Sensor* [24]), and may include in the future sensors that not yet fully been explored. Examples for futuristic sensors include physical-measuring wearable sensors that measure the level of dehydration [25], sensors that measure the level of exposure to sunlight, and sensors that measure weight and height. Examples for futuristic sensors include also psychological-measuring wearable sensors that measure the level of patience, sensors that measure the level of boredom, and sensors that measure the level of fatigue.

Another advantage of the methods described in this paper relates to discovering associations between different sensor measurements, for example, to identify similarities between cardiovascular patterns and oxygen saturation patterns or to discover associations between levels of skin conductivity measured in multiple human body areas.

## VI. CONCLUSIONS AND FUTURE RESEARCH

The paper presents the development of a bio-medical simulator. The simulator is based on:

- A sampling procedure that allows acquiring patterns of 7,871 financial instruments traded in real-time on three

American stock exchanges: *NASDAQ*, *NYSE*, and *AMEX*,

- A classification method to identify similarly behaving pairs, and,
- A user interface that allows observing the identified pairs in real-time.

Positive high correlations were repeatedly observed among pairs that contain patterns that were identified as similar. Classification was based on new machine learning techniques: *self-labeling*, which allows the application of supervised learning methods on unlabeled time-series and *similarity ranking*, which applied on a decision tree learning algorithm to classify time-series regardless of type and quantity. The classification approach proposes to address the anticipated scalability limitation involved with multiple-sensing-based systems as it allows classifying measurements regardless of the type and the quantity of the measurements.

ACKNOWLEDGMENT

The author wishes to show his greatest appreciation to Mr. David Orange, BSEE, JD and LLM (The University of Washington) for his tremendous support and help.